\documentclass[10pt,twocolumn,letterpaper]{article}
\usepackage{cvpr}
\usepackage{comment}
\usepackage{graphicx}
\usepackage{amsmath}
\usepackage{amssymb}
\usepackage{booktabs}
\usepackage{multirow}
\usepackage{enumitem}
\usepackage[pagebackref,breaklinks,colorlinks]{hyperref}
\usepackage{booktabs}

\usepackage[capitalize]{cleveref}
\crefname{section}{Sec.}{Secs.}
\Crefname{section}{Section}{Sections}
\Crefname{table}{Table}{Tables}
\crefname{table}{Tab.}{Tabs.}

\def\confName{CVPR}
\def\confYear{2022}
\def\eg{\emph{e.g}\onedot}

\def\ie{\emph{i.e}\onedot}

\def\etc{\emph{etc}\onedot}

\begin{document}

\title{Trajectory Forecasting from Detection with Uncertainty-Aware Motion Encoding}
\author{Pu Zhang{\small $~^{1}$}, ~Lei Bai{\small $~^{2}$}, ~Jianru Xue{\small $~^{1}$}\thanks{Corresponding author.},~Jianwu Fang{\small $~^{2}$}, ~Nanning Zheng{\small $~^{1}$}, ~Wanli Ouyang{\small $~^{3}$}\\
	\normalsize
	$^{1}$\
Institute of Artificial Intelligence and Robotics, Xi'an Jiaotong University, China\\
	\normalsize
$^{2}$\, Chang'an University, China,
	\normalsize
$^{3}$\, The University of Sydney, Australia\\
	\normalsize
}
\begin{comment}
	\normalsize
	{zhangpu2016}@stu.xjtu.edu.cn,
	\normalsize
	{jrxue,nnzheng}@mail.xjtu.edu.cn,
	\normalsize
	{fangjianwu}@chd.edu.cn,
	\normalsize
    wanli.ouyang@sydney.edu.au
\end{comment}
\maketitle

%%%%%%%%% ABSTRACT
\begin{abstract}
   Trajectory forecasting is critical for autonomous platforms to make safe planning and actions.
   Currently, most trajectory forecasting methods assume that object trajectories have been extracted and directly develop trajectory predictors based on the ground truth trajectories. However, this assumption does not hold in practical situations. Trajectories obtained from object detection and tracking are inevitably noisy, which could cause serious forecasting errors to predictors built on ground truth trajectories.

   In this paper, we propose a trajectory predictor directly based on detection results without relying on explicitly formed trajectories. Different from the traditional methods which encode the motion cue of an agent based on its clearly defined trajectory, we extract the motion information only based on the affinity cues among detection results, in which an affinity-aware state update mechanism is designed to take the uncertainty of association into account. In addition, considering that there could be multiple plausible matching candidates, we aggregate the states of them. This design relaxes the undesirable effect of noisy trajectory obtained from data association.
   Extensive ablation experiments validate the effectiveness of our method and its generalization ability on different detectors.
   Cross-comparison to other forecasting schemes further proves the superiority of our method. Code will be released upon acceptance.

\end{abstract}

%%%%%%%%% BODY TEXT
\section{Introduction}\label{sec:intro}
\begin{figure}[!t]
\begin{center}
\begin{subfigure}{0.45\textwidth}
\includegraphics[width=1\linewidth]{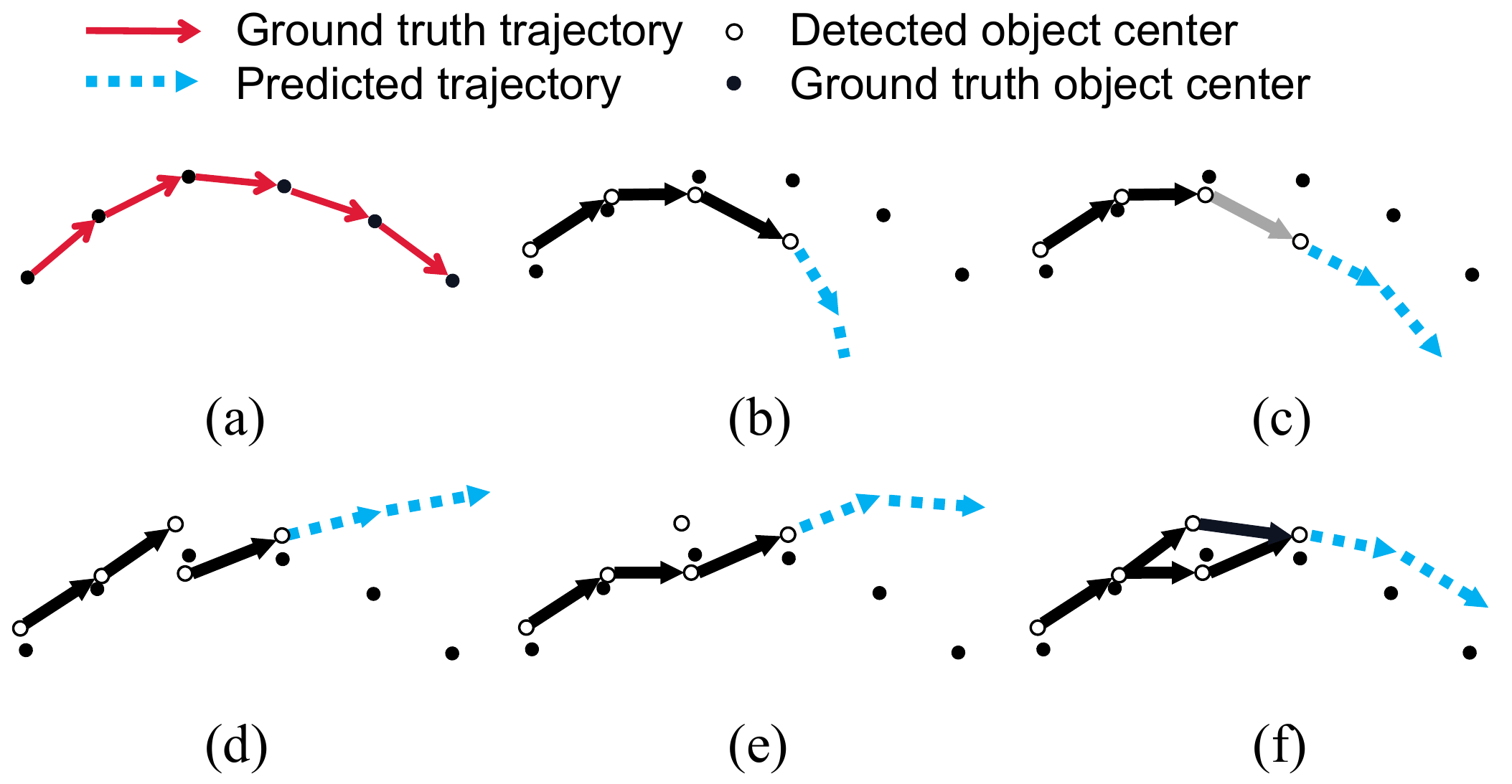}
\end{subfigure}
\caption{Problems of forecasting using trajectories from detectors and trackers. (a) A piece of ground truth trajectory. (b)\&(c): The last detection result is with a large localization error. (b) Completely trusting the association results in large error of the predicted trajectory. (c) Taking the uncertainty of association into consideration (grey arrow) and relying more on previous associations but less on the latest yet unreliable one, a decent future trajectory can be forecast. (d-f): Tracked trajectories are often erroneous because of the detection noise, including (d) breaking into fragments or (e) losing motion tendency. (f) Taking multiple plausible candidates into consideration can alleviate the impact of the trajectory noise.}
\label{fig:motivation}
\vspace{-0.6cm}
\end{center}
\end{figure}

Forecasting trajectories of traffic participants (e.g., vehicles) plays an increasingly important role in developing autonomous platforms.
While there have been a large number of efforts devoted to this field\cite{alahi2016social,zhao2019multi,zhang2019sr,kothari2021interpretable,phan2020covernet,fang2020tpnet,narayanan2021divide,liu2021multimodal,ye2021tpcn,narayanan2021divide}, the vast majority of existing studies are developed under the assumption that perfect trajectories have been extracted.
Thus these models directly employ the ground truth historical trajectories as input and encode them as representations, from which future trajectories are generated by a trajectory decoder. In the rest of this paper, we denote the process of encoding historical information for trajectory forecasting as \emph{motion encoding}.

In real-world applications, the above assumption about using ground truth trajectories does not hold since trajectories are normally estimated by cascaded detectors and trackers. However, the estimated trajectories are corrupted ones due to both detection and tracking noise. Employing them as the input of existing trajectory predictors could lead to severe accumulated errors and pose severe threats to the system safety.
Besides, since existing methods are designed upon ground truth trajectories, they will suffer from the following two prominent problems when using trajectories extracted by detectors and trackers (illustrated in Fig.\ref{fig:motivation}). \par

First, existing forecasting methods totally trust the association results. However, two associated detection results among a tracked trajectory could have varying matching uncertainty reflecting the reliability of an association, which is neglected in existing approaches. For example, the last detection result in Fig.\ref{fig:motivation} (b) has a large localization error which leads to low matching confidence, while completely believing this connection would cause a large prediction error in the future trajectory.

Second, previous works only receive the optimal matching results from the trackers, which form a single tracked trajectory for each individual. However, they neglect the useful information provided by other plausible matching detections which could also be contributory to this trajectory.
In Fig.\ref{fig:motivation}(d), the estimated trajectory could break into fragments due to the mismatch to a False Positive (FP) at the previous time step, which largely loses the valuable historical information. Even when detection results are optimally linked together in Fig. \ref{fig:motivation}(e), the tracked trajectory may still not be enough to encode the motion tendency due to the localization drifts of the detection results.

To address the two problems above, we are motivated to build trajectory predictors \emph{directly from detection results rather than on top of explicitly formed trajectories}. In our design, the uncertain relationship between detections is flexible to model, which can avoid the undesirable influence from the noisy trajectories. Specifically, instead of totally trusting every matched detection, we measure the affinity between detections and use these affinity cues as the representation of association uncertainty, \eg, grey arrow in Fig.\ref{fig:motivation}(c), to guide the update of the motion encoding, which is implemented by \emph{Affinity-aware State Update} (ASU) mechanism.
In addition, features of multiple plausible detection candidates, \eg, two detections at the second last time step in Fig.\ref{fig:motivation}(f), are collected together to enhance the state representation for the motion encoding, which is achieved by  \emph{Multiple State Aggregation} (MSA) mechanism.\par

\begin{figure}[!t]
\begin{center}
\begin{subfigure}{0.47\textwidth}
\includegraphics[width=1\linewidth]{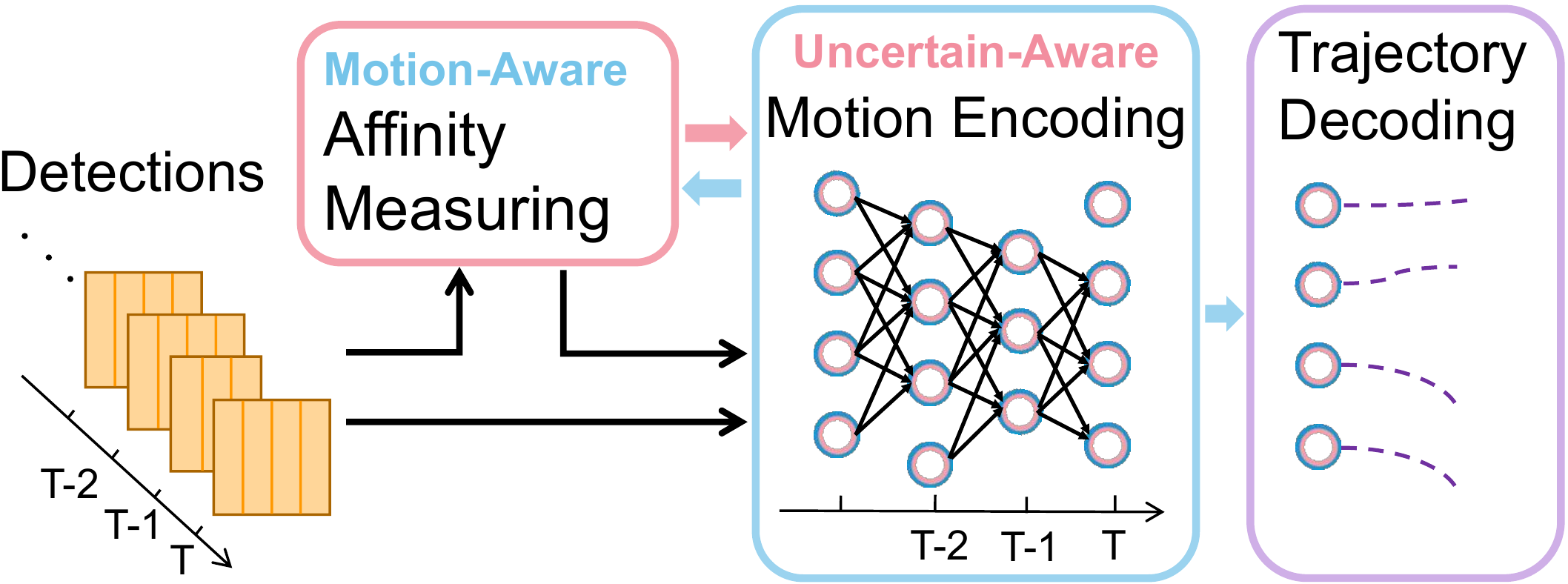}
\end{subfigure}
\caption{
Overview. The affinity measuring module infers the matchness between detections. The motion encoding module generates historical representations. Two modules exchange features over time, which brings the motion-aware and uncertain-aware characters. The encoded motion representation are used by the trajectory decoding module for trajectory forecasting. }%The representation generated from the motion encoding module is used for future trajectory decoding.}
\label{fig:overview}
\vspace{-0.6cm}
\end{center}
\end{figure}

Figure \ref{fig:overview} illustrates the overview of the proposed approach. The proposed model is mainly composed of three modules, motion-aware affinity measuring, uncertain-aware motion encoding, and trajectory decoding. The motion-aware affinity measuring module is designed to
measure the matchness between detection results with long-term motion cues and represent the matching uncertainty. The uncertain-aware motion encoding module encapsulates the proposed ASU and MSA mechanisms to generate more robust historical motion representations using the affinity measures from the affinity measuring module. The trajectory decoding module uses the motion representations from the motion encoding module for predicting future trajectories. Since our model does not generate explicit historical trajectories but forecasts based on the possibly linked historical detections, we call these linked detections across time \emph{implicit trajectories}. Compared with the traditional trajectory forecasting from ground truth trajectories, this setting is based on the practical perception systems, which is more consistent with the deployment stage in real-world environments.

The core contributions are summarized as follows:
\begin{itemize}[noitemsep,topsep=0pt,parsep=0pt,partopsep=0pt]
\item
We propose a trajectory forecasting framework with an uncertain-aware motion encoding process, which does not rely on explicit trajectories but infer the future trajectories directly from observed detections.
\item
We propose an affinity-aware state update mechanism that incorporates the matching uncertainty between detection results into the history encoding process.
\item
We propose a multiple state aggregation mechanism to integrate plausible matching detection candidates into the motion encoding process, which results in more robust historical motion representations.
\end{itemize}

{The proposed ASU and MSA can improve the forecasting results of nonlinear trajectories at 3 seconds for detectors of CenterPoint\cite{yin2021center}, FreeAnchor\cite{zhang2019freeanchor}, SSN\cite{zhu2020ssn}, and PointPillar\cite{lang2019pointpillars} by 3.2, 3.4, 3.8, and 4.1 (\%). We also evaluate our framework on trajectory forecasting designs of MATF\cite{zhao2019multi}, SR-LSTM\cite{zhang2020social}, STGAT\cite{huang2019stgat} based on detection of\cite{yin2021center}, improvements of our designs are 3.5, 3.2 and 2.8 (\%).}
\section{Related Works}

\textbf{Trajectory forecasting.} Studies on trajectory forecasting have investigated various aspects in recent years, \eg, homogeneous\cite{alahi2016social,zhao2019multi,zhang2019sr,pang2021trajectory,kothari2021interpretable,yu2020spatio,shi2021sgcn} or heterogeneous\cite{chandra2019traphic,ma2019trafficpredict,marchetti2020mantra,zheng2021unlimited} social interactions, multimodal modeling\cite{phan2020covernet,fang2020tpnet,salzmann2020trajectron++,narayanan2021divide,liu2021multimodal}, structural map representation\cite{ye2021tpcn,narayanan2021divide}, decision/intention-based\cite{rasouli2019pie,song2020pip,malla2020titan} or goal-based\cite{mangalam2020not,dendorfer2020goal} forecasting, \etc.
While these works improve trajectory forecasting significantly, they all assume that the ground truth  historical trajectories are available. Differently, we focus on improving the robustness of the trajectory forecasting based on practical detectors. Therefore, our proposed motion encoding can be plugged in as the module for encoding motion information for the works above when using them under practical perception situations.
Since the historical information is represented through ASU and MSA, the proposed approach can largely alleviate the impact of noisy detection results or inaccurate historical trajectories.

\textbf{Joint detection and forecasting.}
A few prior works explored the joint solution towards detection and trajectory forecasting\cite{li2020end,casas2020spagnn,
meyer2020laserflow,luo2018fast,casas2018intentnet,zeng2020dsdnet}, in which the detection is conducted on a sequence of frames (\eg, lidar sweeps) and the forecasting is achieved by adding a corresponding forecasting header on each detection proposal.
Different from these works, our work does not depend on the optimization of detection backbones, thus is compatible with different detectors so as to benefit from the rapid development of object detection field, \eg, successful detectors provided in MMDetection3D\cite{mmdetection3d}.

\textbf{Joint tracking and forecasting.}
Recently, a few studies emerged to address the tasks of Multiple Object Tracking (MOT) and trajectory forecasting together~\cite{weng2021ptp,liang2020pnpnet} with shared feature learning, which is similar to the setting of  `forecasting from detection' in this paper. Specifically, the study of \cite{weng2021ptp} employs GNNs to interact the track nodes with the current detection nodes and takes their differences as edge features to learn the affinity matrix for association. It is also under the assumption that trajectories up to the previous time step are given.
The study~\cite{liang2020pnpnet} introduces tracking into the joint detection and forecasting methods, where shared trajectory representation is utilized in both tracking and forecasting modules.
The forecasting approaches~\cite{weng2021ptp,liang2020pnpnet} still rely on explicit tracking results to derive the trajectory representation, which could have performance deterioration when tracking is unstable as discussed in Sec.\ref{sec:intro}. To avoid the undesirable influence from inaccurate tracking, our work is based on detection results only with matching affinity as the implicit trajectory representation and introduces the association uncertainty into the motion encoding process.   \par

\section{Approach}
\textbf{Problem formulation.}
Our target is to forecast trajectories of multiple objects based on detection results of consecutive historical frames. Detection results up to the current time slot $T$ are represented by $\mathcal{D}=\{\mathcal{D}^1,\mathcal{D}^2,...,\mathcal{D}^T\}$, $\mathcal{D}^t=\{d_1^t,d_2^t,...,d_{N_t}^t\}$ is the set of $N_t$ detection results at time slot $t$. Each detection result $d^t_n$,where $n=1,2,...,N_t$, is made of multiple components, we use the following form~\cite{caesar2020nuscenes} as an example,
\begin{equation}
\begin{aligned}\label{eq:det_input}
d_n^t=[d_n^{{\rm{pos}},t},d_n^{{\rm{velo}},t},d_n^{{\rm{size}},t},
d_n^{{\rm{head}},t},d_n^{{\rm{score}},t}],\\
\end{aligned}
\end{equation}
where the elements respectively denote the predicted position $(p_{n,x},p_{n,y})$, velocity $(v_{n,x},v_{n,y})$ \footnote{Movements on the z-axis are ignored.}, size, heading angle, and detection score. The position $d_n^{{\rm{pos}},t}$ is necessary, while others depend on the design of the 3D object detector.

Following the general practice, we use a time window $T_{\rm obs}$ and employ $\mathcal{D}^{T-T_{\rm obs}+1:T}$ to replace $\mathcal {D}^{1:T}$ as the model inputs to predict the future trajectories $\mathcal{F}^T=\{\mathbf{f}_1^T, ... \mathbf{f}_n^T,...,\mathbf{f}_{N_T}^T\}$ of all detections at $T$. Trajectory $\mathbf{f}_n^T$ for the $n$th agent is composed by a sequence of future locations $\mathbf{f}_n^T=\{({p}^t_{n,x}, p^t_{n,y})|t=T+1,T+2,...,T+T_{\rm pred}\}$, where $T_{\rm pred}$ is the time steps of the forecasting horizon.

\begin{figure}[!t]
\begin{center}
\includegraphics[width=0.9\linewidth]{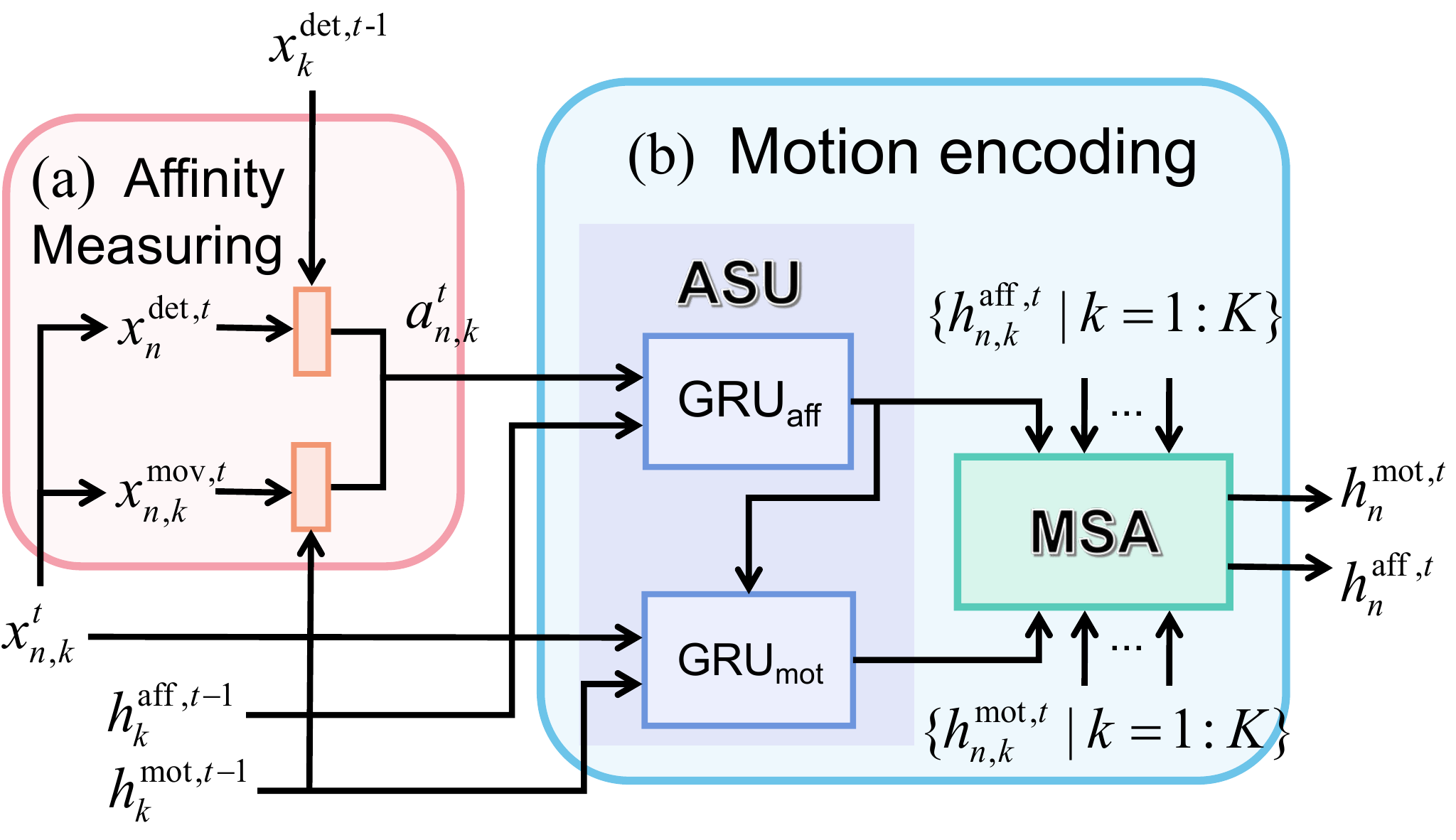}
\caption{Step-wise processing of the affinity measuring and motion encoding modules, % for extracting the historical motion information
where the $k$th detection at $t-1$, \ie $d_k^{t-1}$ is potentially matched with the $n$th detction at $t$, \ie $d_t^{n}$. (a) Two affinity features are calculated for affinity measuring: the short-term feature between $x_k^{{\rm det},t-1}$ and $x_n^{\rm{det},t}$ (Eq.\ref{eq:affinity_det}), the long-term feature drawn from $h_k^{{\rm mot},t-1}$ and the one-step movement $x^{{\rm mov},t}_{n,k}$ (Eq.\ref{eq:affinity_mot}). %, which are summarized as $a_{n,k}^t$.
(b) Towards the motion encoding under noisy circumstances, the $\rm GRU_{aff}$ is introduced in ASU to modulate the affinity features which helps to guide state update for the $\rm GRU_{mot}$ (Eq.\ref{eq:aa-update}). The MSA is deployed to integrate useful information from multiple plausible matched candidates at $t-1$ (Eq.\ref{eq:motion-aggr-fun}). }
\label{fig:overview_step}
\vspace{-0.6cm}
\end{center}
\end{figure}

\subsection{Overview}
As shown in Fig.\ref{fig:overview}, our method takes detection results as the input (detection representation is presented in Sec.\ref{sec:detection}) and obtains motion representation by the following two modules.
The \emph{motion-aware affinity measuring} module (Sec.\ref{sec:affinity}) measures the matchness between detections among every two adjacent frames.
The \emph{uncertainty-aware motion encoding} module (Sec.\ref{sec:motion-encoding}) extracts compact representation for the historical motion information by  affinity-aware state update (Sec.\ref{sec:ASU}) and multiple state aggregation (Sec.\ref{sec:MSA}).
The step-wise processing of the two modules from $t-1$ to $t$ is illustrated in Fig. \ref{fig:overview_step}.

Representation from the motion encoding module is finally employed by the trajectory decoding module to predict future trajectories (Sec.\ref{sec:social}).

\subsection{Detection Representation}\label{sec:detection}
To facilitate the description, we use $m$ to denote the index of any detection at $t-1$ in Sec.\ref{sec:detection} and Sec.\ref{sec:affinity}, while use $k$ for those potentially matched candidates.

We represent the $d_n^t$ (with a predecessor $d_m^{t-1}$) as a feature $x_{n,m}^t$, which is concatenated by two parts,
\begin{equation}
\begin{aligned}\label{eq:det_x}
x_{n,m}^{t}&=[x_{n}^{{\rm{det}},t};x_{n,m}^{{\rm{mov}},t}].\\
\end{aligned}
\end{equation}
One is the summary of the detected object state of $d_n^t$,
\begin{equation}
\begin{aligned}\label{eq:det_feat}
x_{n}^{{\rm{det}},t}&={\rm{MLP_{fus}}}([x_n^{{\rm{velo}},t};x_{n}^{{\rm{size}},t}
x_{n}^{{\rm{head}},t};x_{n}^{{\rm{score}},t}])\\
\end{aligned}
\end{equation} containing all unary information of the detected results in Eq.\ref{eq:det_input} except $d_t^{{\rm pos},t}$, where each input item is embedded through
$x_{}^{*,t}={\rm MLP_*}(d^{*,t})$.
The other, $x_{n,m}^{{\rm{mov}},t}={\rm MLP_{mov}}(d^{{\rm{pos}},t}_{n}-d^{{\rm{pos}},t-1}_{m})$, suggests the movement between $d^{t-1}_m$ and $d^t_n$.
The design of the detection representation shows compatibility to serve as the input for the affinity measuring module, \ie, $x_{n}^{{\rm{det}},t}$ and $x_{n,m}^{{\rm{mov}},t}$ are respectively used in Eq.\ref{eq:affinity_det} and Eq.\ref{eq:affinity_mot} for calculating the short-term and long-term affinity features.
In addition, using movement can make the motion modeling not to depend on the global coordinates.

\subsection{Motion-Aware Affinity Measuring}\label{sec:affinity}
Similar to the tracking-by-detection MOT framework, an affinity network is deployed to learn a similarity function between any two detected instances.
Inspired by the studies which leverage recurrent networks to strengthen the capability of affinity from multiple frames computing\cite{milan2017online,fang2018recurrent,sadeghian2017tracking,kim2018multi,kim2021discriminative}, we introduce an affinity network with the long-term motion-aware features as input, named motion-aware affinity measuring, supervised by the matching as well as trajectory forecasting losses.
Specifically, two kinds of features are served as the affinity measuring inputs, one is the short-term feature representing the similarity of detected object states, the other is the long-term feature which describes the quality of the agent dynamics.  \par

\textbf{Short-term affinity feature.}
We first calculate the correlation between two detection representations through an absolute subtraction~\cite{zhang2019robust},
\begin{equation}
\begin{aligned}\label{eq:affinity_det}
a_{n,m}^{{\rm{det}},t}&=|x_{n}^{{\rm{det}},t}-x_{m}^{{\rm{det}},t-1}|.
\end{aligned}
\end{equation}\par
\textbf{Long-term affinity feature.}
To make the affinity measure sensitive to dynamics of moving objects, we introduce
\begin{equation}
\begin{aligned}\label{eq:affinity_mot}
&a_{n,m}^{{\rm{mot}},t}={\rm MLP_{mot}}([x_{n,m}^{{\rm mov},t};h_{m}^{{\rm{mot}},t-1}])\\
\end{aligned}
\end{equation}
to represent the correlation between the historical motion and the current movement,
where $h_{m}^{{\rm{mot}},t-1}$ is from the hidden state of $\rm GRU_{mot}$ at $t-1$. $\rm GRU_{mot}$ is a gated recurrent unit used for the motion encoding (introduced in Sec.\ref{sec:motion-encoding}), hidden states of which provide strong motion cues of the objects location at the next time step. Thus, the combination of $h_{m}^{{\rm{mot}},t-1}$ and $x_{n,m}^{{\rm mov},t}$ can describe whether the oncoming connection is coherent with the historical motion. \par
The final affinity feature is expressed by
\begin{equation}
\begin{aligned}\label{eq:affinity_feat}
a_{n,m}^t&=[a_{n,m}^{{\rm{mot}},t};a_{n,m}^{{\rm{det}},t}].
\end{aligned}
\end{equation}
Based on the affinity feature, we calculate an affinity matrix $\mathbf{S}^t$, the element (affinity score) of which is generated by
\begin{equation}
\begin{aligned}\label{eq:affinity_score}
s_{n,m}^{t}=\sigma ({\rm MLP_{aff}}(a_{n,m}^{t})),
\end{aligned}
\end{equation}
where $\sigma$ is the sigmoid function.  The affinity score computed by ${\rm MLP_{aff}}$ is served as the confidence that $d_m^{t-1}$ is associated with $d_n^{t}$. \par
The function of the affinity measuring is two-fold in our designs. First, the affinity feature indicating the feature-wise matching uncertainty provides cues for the affinity-aware state update (Eq.\ref{eq:aff-GRU},\ref{eq:aa-update}). Second, the affinity matrix is served to guide the state aggregation process (Eq.\ref{eq:motion-aggr-fun},\ref{eq:selection}). \par

\subsection{Uncertain-Aware Motion Encoding}\label{sec:motion-encoding}
With the $\mathbf{S}^{t}$ from the affinity measuring network, $K$ candidates with top-$K$ affinity scores are selected. These $K$ candidates at $t-1$ can potentially match with a detection instance at the current time $t$. In the following, we use the subscript $k$ to denote the $k$th matched candidate and demonstrate the detailed state update process from $t-1$ to $t$ in the motion encoding module.
\vspace{-0.1cm}
\subsubsection{Basic Motion Encoding}
First, we give a basic implementation of the motion encoding to manifest our uncertain-aware designs. Given $d_k^{t-1}$, a matched detection with the $d_n^{t}$, the {basic motion encoding network} will update the representation of this implicit trajectory by:
\begin{equation}
\begin{aligned}\label{eq:basic-motion-encoding}
h_{n, k}^{{\rm{mot}},t}&={\rm{GRU_{mot}}}(x^{t}_{n,k},h^{{\rm{mot}},t-1}_k),\\
\end{aligned}
\end{equation}
where $x^{t}_{n,k}$ is given by Eq.\ref{eq:det_x}, $h^{{\rm{mot}},t-1}_k$ is the trajectory representation at the previous step, which is updated as $h^{{\rm{mot}},t}_{n,k}$ through the gated recurrent unit ${\rm GRU_{mot}}$.
\vspace{-0.1cm}
\subsubsection{Affinity-Aware State Update}\label{sec:ASU}
To incorporate the association uncertainty cues into the motion encoding, we revise the state update in Eq.\ref{eq:basic-motion-encoding} to
%the state update function shown in Eq.\ref{eq:basic-motion-encoding} to
\begin{equation}
\begin{aligned}\label{eq:basic-aa-update}
h_{n,k}^{{\rm{mot}},t}&={\rm{GRU_{mot}}}([x^{t}_{n,k},u^{t}_{n,k}],h^{{\rm{mot}},t-1}_k),\\
\end{aligned}
\end{equation}
where, $u^t_{n,k}$ is expected to describe the uncertainty about whether $d_k^{t-1}$ is associated with $d_n^{t}$. \par

Here, a straightforward way of the design of $u^t_{n,k}$ is directly using the affinity score $s^{t}_{n,k}$ (Eq.\ref{eq:affinity_score}), \ie $u^t_{n,k}=s^{t}_{n,k}$.
However, $s^{t}_{n,k}$ is an overall confidence, which just has limited information. A better choice is to use the multi-dimensional affinity feature (Eq.\ref{eq:affinity_feat}), \ie, $u^t_{n,k}=a^{t}_{n,k}$, which contains richer information than the scalar $s^{t}_{n,k}$.
Assuming that an implicit trajectory of length $L$ is formed after $L$ time steps (with $d_n^t$ as the last node), the affinity features belonging to this implicit trajectory result in another sequence of length $L-1$. We further introduce another GRU, $\rm GRU_{aff}$, to modulate this chain of affinity features,
\begin{equation}
\begin{aligned}\label{eq:aff-GRU}
h_{n,k}^{{\rm{aff}},t}&={\rm{GRU_{aff}}}(a^{t}_{n,k},h^{{{\rm{aff}},t-1}}_{k}).\\
\end{aligned}
\end{equation}
Compared with $a^{t}_{n,k}$, hidden state of $\rm GRU_{aff}$ maintains a time-related uncertain feature, which is a better choice of $u_{n,k}^{t}$.
Equation \ref{eq:basic-aa-update} is finally implemented by
\begin{equation}
\begin{aligned}\label{eq:aa-update}
h_{n,k}^{{\rm{mot}},t}&={\rm{GRU_{mot}}}([x^{t}_{n,k},h_{n,k}^{{\rm{aff}},t}],h^{{\rm{mot}},t-1}_k).\\
\end{aligned}
\end{equation}

\subsubsection{Multiple State Aggregation}\label{sec:MSA}
As discussed in Sec.\ref{sec:intro}, there could be multiple detection candidates contributed to a same trajectory due to the detection noise.
Therefore, we introduce an aggregation module to integrate the features of multiple plausible candidates. Specifically,
hidden states of $K$ matched candidates are aggregated as follows:
\begin{equation}
\begin{aligned}\label{eq:motion-aggr-fun}
h_{n}^{{\rm{mot}},t}&=\sum\nolimits_{k=1}^{K} \alpha_{n,k}\cdot (g^{\rm{mot}}_{n,k}\odot h_{n,k}^{{\rm{mot}},t}),\\
\end{aligned}
\end{equation}
where $\odot$ denotes the element-wise product operation, $\alpha_{n,k}$ together with $g^{\rm mot}_{n,k}$ are served as the feature selection to extract useful information from the matched candidates,
\begin{equation}
\begin{aligned}\label{eq:selection}
\alpha_{n,k}^t&=\frac{e^{s^{t,{\rm{lg}}}_{n,k}}}{\sum_{l=1}^{K}{e^{s^{t,{\rm{lg}}}_{n,l}}}},s^{t,{\rm{lg}}}_{n,k}=\sigma^{-1}(s^t_{n,k}),\\
g_{n,k}^{\rm mot}&=\sigma(W^{\rm lay}[h_{n,k}^{{\rm{mot}},t};h_{n,k}^{{\rm{mot}},t-1};x_{n,k}^{t}]+b^{\rm lay}),\\
\end{aligned}
\end{equation}
where $\sigma^{-1}$ denotes the logit function, $W^{\rm lay}$ and $b^{\rm lay}$ are parameters of a linear layer. $\alpha_{n,k}$ is from the affinity scores (Eq.\ref{eq:affinity_score}), which is designed to restrain the contribution of candidates having low affinities and keep the ones with high affinities. The $g^{\rm mot}_{n,k}$ is used to select different features from a certain candidate.
The aggregated feature is prepared as the hidden feature of $\rm GRU_{mot}$ for the next time step.

To receive the hidden states of $\rm GRU_{aff}$, another aggregation function is introduced with the same form of Eq.\ref{eq:motion-aggr-fun},
\begin{equation}
\begin{aligned}\label{eq:affinity-aggr}
h_{n}^{{\rm{aff}},t}&=\sum\nolimits_{k=1}^{K} \alpha_{n,k}\cdot (g^{\rm{aff}}_{n,k}\odot h_{n,k}^{{\rm{aff}},t}),\\
g_{n,k}^{\rm aff}&=\delta(W^{\rm aff}[h_{n,k}^{{\rm{aff}},t};h_{n,k}^{{\rm{aff}},t-1};a_{n,k}^{t}]+b^{\rm aff}),\\
\end{aligned}
\end{equation}
where $W^{\rm aff}$ and $b^{\rm aff}$ are learned parameters,
$\alpha_{n,k}$ is shared with the one in Eq.\ref{eq:motion-aggr-fun}. Equation \ref{eq:affinity-aggr} can be considered as making a summary of multiple uncertainty cues ($h_{n,k}^{{\rm{aff}},t}$ with $k=1:K$), which is prepared as the updated hidden state for $\rm GRU_{aff}$ input at the next step.\par
\textbf{Discussion of the MSA design.}
Similar to Multiple Hypothesis Tracking (MHT) based methods \cite{blackman2004multiple, kim2015multiple,kim2018multi}, MSA retains information of multiple candidates. Instead of storing hypothetical trajectories in MHT, MSA updates the trajectory representation by aggregating features of multiple candidates
so that the required memory does not change over time.
Different from the multi-track pooling in \cite{kim2021discriminative}, where features of the non-target tracks are pooled together to make the affinity computing more discriminative,
we aggregate features of plausible matched candidates, since they could be different observations of one object, thus to result in more robust history representations towards forecasting.

\subsection{Trajectory Decoding}\label{sec:social}
After the motion encoding for $T_{obs}$ time steps up to $t = T$, representation $h^{{\rm mot},T}_{n}$ for the $n$th individual is achieved. The proposed model can be further combined with existing Social Interaction Modeling (SIM) functions which converts the representation $h^{\rm{mot},T}_{n}$ of the $n$th individual to $p_n^T$ with social-aware effects. In the experiments, we adopt the methods in~\cite{zhang2019sr,huang2019stgat,zhao2019multi} for implementing the SIM function.
Given the features $p_n^{T}$ from the motion encoding and SIM, the future trajectories are generated through $\mathbf{f}_{n}^{T}={\rm MLP_{dec}}(p_{n}^T)$.
\subsection{Optimization}\label{sec:loss}
The loss functions are composed of two parts,
\begin{equation}
\begin{aligned}\label{eq:loss}
l=l_{\rm{traj}}+\lambda {\sum\nolimits_{t=1}^{T_{\rm obs}}{l}_{\rm{aff}}^t}/(T_{\rm obs}-1),
\end{aligned}
\end{equation}
where $\lambda$ is a coefficient to balance the importance of the two parts.
The first part, $l_{\rm traj}$, is the smooth $\rm L1$ loss on the trajectory output at $t=T$, which is widely used in many trajectory forecasting methods~\cite{liang2020pnpnet,casas2018intentnet,gu2021densetnt}.
The second part,  $l^t_{\rm aff}$, is a binary cross entropy loss on the affinity scores (Eq.\ref{eq:affinity_score}) of detection results at each two adjacent frames,
\begin{equation}
\begin{aligned}\label{eq:aff-loss}
l_{\rm{aff}}^t=-\frac{1}{MN}\sum\nolimits_{n=1}^{N}\sum\nolimits_{m=1}^{M}\hat{s}_{n,m}^{t}\log s_{n,m}^{t}\\+(1-\hat{s}_{n,m}^{t})\log (1-s_{n,m}^{t}),
\end{aligned}
\end{equation}
{where $\hat{s}_{n,m}^{t}\in \{0, 1\}$ denotes whether the two detections, $d_n^t$ and $d_m^{t-1}$, belong to a same trajectory. To generate $\hat{s}_{n,m}^{t}$, we follow the labeling process in~\cite{zhang2019robust}, where the global ID of detections are first generated through matching with the GT boxes at 0.5 IoU, the binary label for a pair of detection results is set to 1 if they are with the same ID.}

\section{Experiments}
The proposed approach is ablated and analyzed in Sec.\ref{sec:ablation} and Sec.\ref{sec:analysis} mainly
based on the the detector of CenterPoint~\cite{yin2021center} \footnote{We use the model \texttt{centerpoint\_voxel\_1440\_dcn(flip)} provided in \cite{centerpointgit}.}. We also evaluate our model on other three detectors to evaluate its generalization ability. In Sec.\ref{sec:comparison}, we compare the proposed approach with other competitive forecasting methods, including the joint tracking and forecasting methods~\cite{weng2021ptp,liang2020pnpnet}, as well as the ones towards joint detection and forecasting\cite{casas2020spagnn,li2020end,meyer2020laserflow,zeng2020dsdnet,zeng2019end,sadeghian2018car}.

\subsection{Datasets and Metrics}

\textbf{Datasets.} We validate the proposed model on nuScenes~\cite{caesar2020nuscenes}, a self-driving dataset with 1000 20-second sequences sampled at a frame rate of 20 Hz. Annotations of 3D objects are given at 2Hz. The dataset is augmented to 20Hz based on the official toolkit. We follow the official train/val split and use the vehicle category for experiments in Sec.\ref{sec:ablation} and Sec.\ref{sec:analysis}, \ie, objects annotated as car, trailer, truck, bus, and construction\_vehicle, with the assumption that the fine-grained class label is unknown. The models are expected to predict the future trajectories for 3s with a time interval of 0.5s, based on 2s historical data. \par

\textbf{Metrics.} The following two metrics are used in the experimental results. First, we use the standard metric, Final Displacement Error (FDE) at 3s for accuracy evaluation (fde@3s), where prediction metrics are computed on True Positive (TP) detections with a recall rate at 0.6 and 0.8. This metric is used in the recent studies focused on joint detection and forecasting~\cite{casas2020spagnn,li2020end}. Second, we also evaluate on nonlinear trajectories using the methodologies in \cite{zhang2020social}(nl\_fde@3s), where the future ground truth trajectories are fitted through the least squares polynomial fitting. The sum of the fitting residuals is used to represent the nonlinearity degree. Nonlinear samples are selected with the fitting residual larger than 0.1.

\subsection{Implementation Details}\label{sec:detail}
During training, we collect detection results from $B$ different sequences, where each sequence starts from a random time step $t$ and ends at $t+T_{obs}-1$ and with the maximum $N$ detection results in each frame, which results in an input batch with the size of $[B, N, T_{obs},...]$. We set $B=128, N=100$, and $T_{obs}=20$. For ground truth trajectories shorter than 3s, we use all valid fragments for training. For training the affinity net, we consider the detection pairs within the distance of $\theta_d=10$ to limit the memory use. The maximum number of matched candidates $K$ is set to 10.
The hidden dimension of the $\rm GRU_{mot}$ and $\rm GRU_{aff}$ are set to 64.
The dimension of the $x_n^{\rm det}$, $x_{n,m}^{\rm mov}$ in Eq.\ref{eq:det_x}, and the long-term affinity feature $a_{n,m}^{\rm mot}$ in Eq.\ref{eq:affinity_mot} are set to 64, 32 and 64 respectively.
We use Adam optimizer to train for 20 epochs with an initial learning rate of 0.003, and decay the learning rate 6 times by a decay factor of 0.6. The weight $\lambda$ is dynamically decreased from 1.0 to 0.1 in half of all training epochs. Data batches are randomly rotated and flipped during training for data augmentation.

\subsection{Ablation Study}\label{sec:ablation}
We analyze the components of the proposed method in this section based on detection results of CenterPoint\cite{yin2021center}.
\vspace{-0.2cm}
%We conduct ablation analysis on components of the proposed method in this section. The detection results used in this section are from the CenterPoint detector\cite{yin2021center}.\par
\subsubsection{Designs in the Baseline Models}
\textbf{Training on clean/noise data.}
The model 0\&1 in Tab.\ref{table:baseline} are used for investigating the effect of training using ground truth trajectories or the trajectories from detection and tracking. Both of them employ the concise tracking implemented in\cite{yin2021center} to associate the detection results in test phase. The difference is that the model 0 is trained on the ground truth historical trajectories, while the model 1 is trained on the online associated trajectories generated through the same way as the test phase.
As can be observed, the performance of the model 0 is much worse than other baselines, which confirms the huge gap between ground truth trajectories in the training phase and trajectories from trackers in the test phase.
%drops sharply when test based on practical detectors.
The model 1 largely attenuates this negative effect by training with tracked trajectories from detection results.\par
\textbf{Motion-aware affinity measuring.} Based on the model 1, we investigate different implementations of the affinity measuring module for model 2\&3. Both of them employ the affinity measure to replace the tracking\cite{yin2021center} in model 0\&1.
The model 2 only uses the short-term affinity feature (Eq.\ref{eq:affinity_det}), while the model 3 employs the motion-aware affinity measuring which exploits both of the short-term and the long-term affinity feature (Eq.\ref{eq:affinity_feat}).
\begin{table}[tbp]
\centering
\small
\begin{tabular}{l|ll|l|l}
\toprule[1pt]
\multirow{2}{*}{ID} &
  \multicolumn{1}{c}{\multirow{2}{*}{Det.}} &
  \multicolumn{1}{c|}{\multirow{2}{*}{Aff.}} &
  \multicolumn{1}{c|}{fde@3s} &
  \multicolumn{1}{c}{nl\_fde@3s} \\
  \multicolumn{1}{c}{} &
  \multicolumn{1}{|c}{} &
  \multicolumn{1}{c|}{} &
  \multicolumn{1}{c|}{rec@0.6 / 0.8}&
  \multicolumn{1}{c}{rec@0.6 / 0.8}\\
  \midrule[0.5pt]
0 &$\times$ &Dist.& 148.8 / 156.3  &359.9 / 375.6 \\
1 &\checkmark& Dist.      &118.5 / 124.9  &290.6 / 309.6   \\
2 &\checkmark& S-Aff.    &117.5 / 122.6  & 286.3 / 301.1 \\
3 &\checkmark& M-Aff.     &115.0 / 120.0  & 281.7 / 296.3  \\
\bottomrule[1pt]
\end{tabular}
\caption{Baseline models. {Det.}: trained from online tracked detections. {Aff.}: calculating the affinity among detections by Euclidead Distance (Dist.), learned short-term affinity (S-Aff.), or our motion-aware affinity (M-Aff.).
}\label{table:baseline}
\end{table}
By introducing the affinity network, the model 2\&3 jointly optimize the two tasks on the given detections, \ie, matching and forecasting, and further improve the performance. Thanks to our motion-aware affinity measure design which incorporates long-term motion cues in association, the model 3 outperforms the model 2 by around 2\% on fde. Thus, we regard the model 3 as a strong baseline in the following sections to manifest the proposed components playing with the corrupt data.
\vspace{-0.2cm}
\subsubsection{Main Components}
Based on the model 3 in Tab.\ref{table:baseline}, we then analyze the effect of two main components, the multiple state aggregation ({MSA} in Sec.\ref{sec:MSA}) and the affinity-aware state update ({ASU} in Sec.\ref{sec:ASU}). The results are given in Tab.\ref{table:ablation}.
For models without MSA, only a single candidate with the highest affinity is kept, thus the feature $h_{n,k}^{{\rm{mot}},t}$ obtained from Eq.\ref{eq:aa-update} is directly used for the trajectory decoder input. Models without ASU do not have the uncertainty input item, \ie, {$u_{n,k}^t$ in Eq.\ref{eq:basic-aa-update}}.\par
\begin{table}[h!]
\centering
\small
\begin{tabular}{l|ll|l|l}
\toprule[1pt]
\multirow{2}{*}{ID} &
  \multicolumn{1}{c}{\multirow{2}{*}{ASU}} &
  \multicolumn{1}{c|}{\multirow{2}{*}{MSA}} &
  \multicolumn{1}{c|}{fde@3s} &
  \multicolumn{1}{c}{nl\_fde@3s} \\
  \multicolumn{1}{c}{} &
  \multicolumn{1}{|c}{} &
  \multicolumn{1}{c|}{} &
  \multicolumn{1}{c|}{rec@0.6 / 0.8}&
  \multicolumn{1}{c}{rec@0.6 / 0.8}\\
  \midrule[0.5pt]
3 &  &    &115.0 / 120.0    &281.7 / 296.3  \\
4 &\checkmark  &    &111.9 / 116.8     &272.4 / 286.6  \\
5 &    &\checkmark  &112.8 / 116.6    &273.2 / 284.5  \\
6 &\checkmark    &\checkmark  &111.0 / 114.8    &269.3 / 280.1 \\
\bottomrule[1pt]
\end{tabular}
\caption{Component analysis of MSA and ASU.}
\label{table:ablation}
\vspace{-0.2cm}
\end{table}
We use ()/()\% to denote the improvement percentage on nl\_fde at the recall rate of 0.6/0.8.
With MSA, the model 5 outperforms the model 3 by 3.0/4.0\%. The soft aggregation can benefit from information of multiple plausible candidates, thus alleviating the impact of the incorrect association on motion modeling, leading to more robust predicting results.
When model 3 and model 4 are compared, the performance gains of ASU are 3.3/3.3\%. As the ASU mechanism uses the affinity features and modulates them through $\rm GRU_{aff}$, the resulting feature can express the uncertainty cues about how one detection matched with its predecessors, which is useful for deriving representations for trajectory forecasting. The overall improvement from the two components in model 6 are 4.4/5.5\% relative to model 3.

\subsection{Analysis}\label{sec:analysis}
\textbf{Uncertainty feature used in ASU.}
According to Sec.\ref{sec:ASU}, there are three options for the uncertainty input items $u_{n,k}^t$ in Eq.\ref{eq:basic-aa-update}: affinity score (Eq.\ref{eq:affinity_score}), affinity feature (Eq.\ref{eq:affinity_feat}) and hidden states from $\rm GRU_{aff}$ (Eq.\ref{eq:aff-GRU}).
The experimental results are given in Tab.\ref{table:abla_afeat}.\par
\begin{table}[ht!]
\small
\centering
\begin{tabular}{l|l|l|l}
\toprule[1pt]
\multirow{2}{*}{ID} &
  \multicolumn{1}{c|}{ASU} &
  \multicolumn{1}{c|}{fde@3s} &
  \multicolumn{1}{c}{nl\_fde@3s} \\
  \multicolumn{1}{c|}{} &
  \multicolumn{1}{c|}{feat.use} &
  \multicolumn{1}{c|}{rec@0.6 / 0.8}&
  \multicolumn{1}{c}{rec@0.6 / 0.8}\\
  \midrule[0.5pt]
3  &- &115.0 / 120.0   &281.7 /296.3  \\
4-a  &affinity score &114.1 / 119.6   &279.5 / 295.3  \\
4-b  &affinity feature &113.2 / 118.1   &276.0 / 290.8 \\
4  &$\rm GRU_{aff}$ feature &111.9 / 116.8   &272.4 / 286.6  \\
%2-b &\checkmark& \checkmark&  &  &10&  &107.4  &110.2  &267.7  &275.6  \\
\bottomrule[1pt]
\end{tabular}
\caption{Different design choices of the affinity-aware feature.}
\label{table:abla_afeat}
\end{table}
In Tab.\ref{table:abla_afeat}, the affinity score (model 4-a) can be helpful, but it is not expressive enough to describe the matching situation of two detection results, which limits its potential in forecasting. In comparison, the multi-dimensional affinity feature (model 4-b) provides more information of association uncertainty which fits better into the forecasting task. Since affinity feature (model 4-b) is only responsible for the association between $t-1$ and $t$, $\rm GRU_{aff}$ (model 4) is introduced to modulate historical affinity features which can encapsulate more comprehensive uncertainty information of the implicit trajectory and result in better performance.

\textbf{Social interaction model.}
The proposed model can be combined with social interaction models.
We use SIMs in three representative predictors, \ie, MATF\cite{zhao2019multi}, SR-LSTM\cite{zhang2020social}, and STGAT\cite{huang2019stgat}, to improve the model capacities. Slight modifications are made to match our designs, \eg trajectory encoders are replaced by our uncertain-aware motion encoding.
The results are given in Tab.\ref{table:SIM}. We conduct experiments for both model 3 (baseline without our new components) and model 6 (our final model) to evaluate the effectiveness of our uncertain-aware motion encoding.
As can be observed, models of 6-a,b,c obviously outperform models of 3-a,b,c, which shows the excellent generalization ability of our designs.\par
\textbf{Test on different detectors.} We also conduct experiments based on detection results from different detectors. The models and weights of these detectors come from mmDetection3D\cite{mmdetection3d}. The results in Tab.\ref{table:detectors} demonstrate the outstanding generalization ability of the proposed model on different detectors. The averaged improvement of the proposed components are around {3.3\%-3.9\%}.\par
\begin{figure}[!h]
\begin{center}
\begin{subfigure}{0.43\textwidth}
\includegraphics[width=1\linewidth]{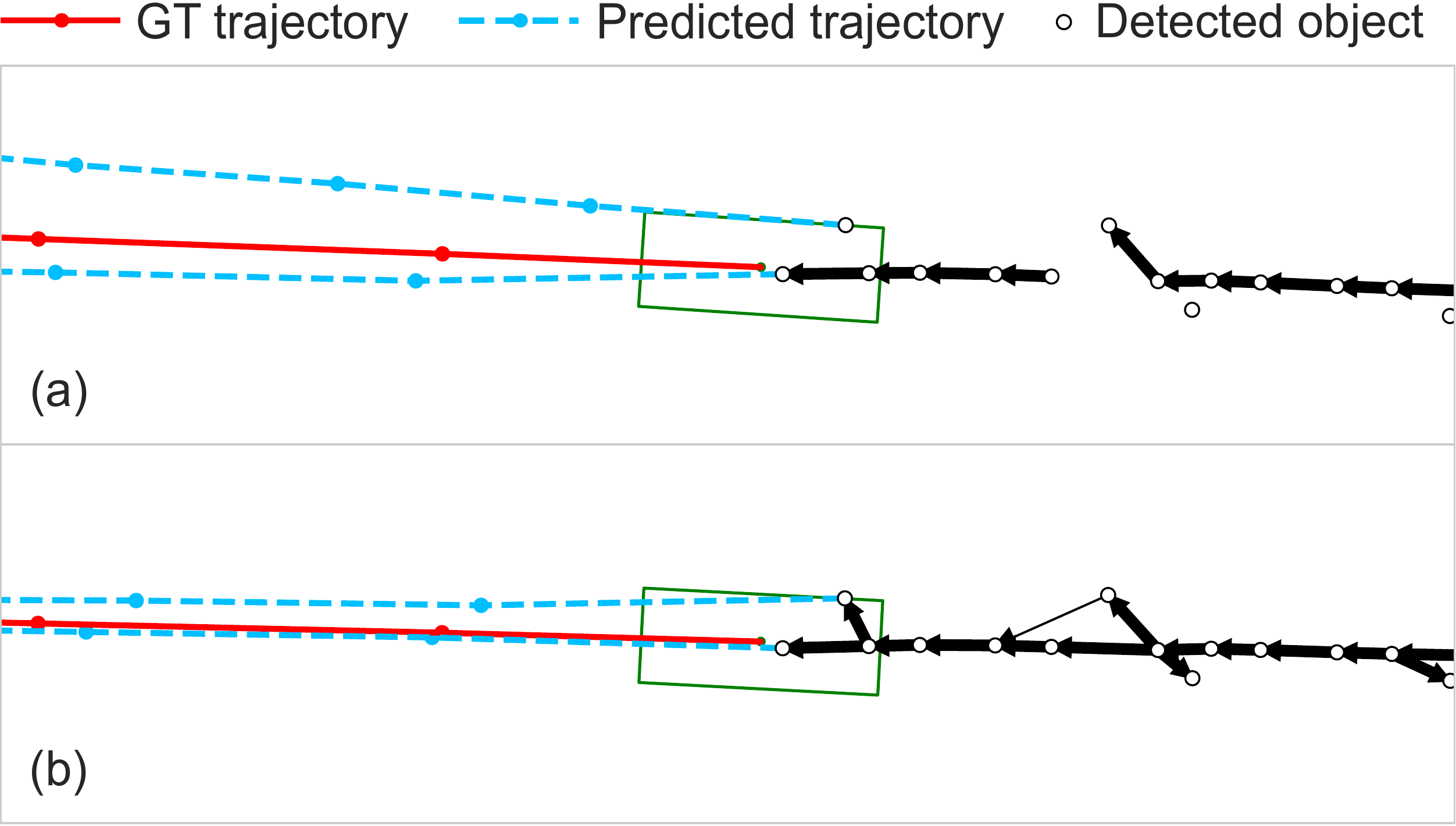}
\end{subfigure}
\caption{Forecasting results without (a) / with (b) the proposed mechanisms of ASU and MSA.
Black arrows represent the implicit trajectory, thickness of which reflects the selecting weight $\alpha$ in Eq.\ref{eq:selection}. Multiple predicted trajectories denote there could be FPs appeared around the object.}
\label{fig:qualitative}
\vspace{-0.6cm}
\end{center}
\end{figure}

\begin{table}[]
\centering
\small
\begin{tabular}{l|l|l|l}
\toprule[1pt]
\multirow{2}{*}{ID} &
  \multicolumn{1}{c|}{\multirow{2}{*}{SIM}} &
  \multicolumn{1}{c|}{fde@3s}&
  \multicolumn{1}{c}{nl\_fde@3s} \\
 &&
  \multicolumn{1}{c|}{rec@0.6/0.8} &
  \multicolumn{1}{c}{rec@0.6/0.8}\\
  \midrule[0.5pt]
3  &\multirow{2}{*}{-} &115.0 / 120.0 &281.7 / 296.3 \\
6  & &111.0 / 114.8 &269.3  / 280.1 \\
\midrule[0.5pt]
3-a  &\multirow{2}{*}{MATF\cite{zhao2019multi}} &108.9 / 112.7  &263.7 / 274.3 \\
6-a  & &105.9 / 110.0  &254.6 / 266.3\\
\midrule[0.5pt]
3-b  &\multirow{2}{*}{SR-LSTM\cite{zhang2020social}} &108.0 / 111.3 &259.4 / 269.3 \\
%6-b  & &\textcolor{red}{105.0 / 108.9}  &\textcolor{red}{249.1 / 260.3} \\
6-b  & &{104.9 / 108.6}  &{251.1 / 261.7} \\
\midrule[0.5pt]
3-c  &\multirow{2}{*}{STGAT\cite{huang2019stgat}} &114.4 / 118.3 &275.7 / 287.5\\
6-c  & &110.6 / 114.6  &268.1 / 279.9 \\
\midrule[0.5pt]
\midrule[0.5pt]
&{Avg. gains} &{3.1 / 3.1 (\%)} & 3.5 / 3.5 (\%)  \\
%2-b &\checkmark& \checkmark&  &  &10&  &107.4  &110.2  &267.7  &275.6  \\

\bottomrule[1pt]
\end{tabular}
\caption{%Trajectory forecasting
Performance with different social interaction models based on the detection results from CenterPoint\cite{yin2021center}.}
\label{table:SIM}
\end{table}
\begin{table}[]
\small
\centering
\begin{tabular}{l|l|l|l}
\toprule[1pt]
\multirow{2}{*}{ID} &
  \multicolumn{1}{c|}{\multirow{2}{*}{Detector}} &
  \multicolumn{1}{c|}{fde@3s}&
  \multicolumn{1}{c}{nl\_fde@3s} \\
 &&
  \multicolumn{1}{c|}{rec@0.6/0.8} &
  \multicolumn{1}{c}{rec@0.6/0.8}\\
  \midrule[0.5pt]
3-b  &\multirow{2}{*}{CenterPoint\cite{yin2021center}} &108.0 / 111.3 &259.4 / 269.3\\
%6-b  & &105.1 / 108.6 &251.6 / 262.0\\
%6-b  & &\textcolor{red}{105.0 / 108.9}  &\textcolor{red}{249.1 / 260.3} \\
6-b  & &{104.9 / 108.6}  &{251.1 / 261.7} \\
\midrule[0.5pt]
3-b  &\multirow{2}{*}{FreeAnchor\cite{zhang2019freeanchor}} &107.5 / 120.3&266.6 / 292.2\\
6-b  & &104.1 / 114.9 &257.6 / 277.8\\
  \midrule[0.5pt]
%2-b  &\multirow{2}{*}{SSN\cite{zhu2020ssn}} &118.0  &121.3 &279.8    &293.6  \\
%5-b  & &113.8  &116.8 &269.3    &282.2\\
3-b  &\multirow{2}{*}{SSN\cite{zhu2020ssn}} &118.0 / - &279.8 / -  \\
6-b  & &113.8 / - &269.3 / -\\
  \midrule[0.5pt]

3-b  &\multirow{2}{*}{PointPillar\cite{lang2019pointpillars}} &123.0 / - &293.6    / -\\
6-b  & &118.6 / - &281.7 / -  \\
\midrule[0.5pt]
\midrule[0.5pt]
&{Avg. gains} &{3.3 / 3.5 (\%)} & 3.6 / 3.9 (\%)  \\
\bottomrule[1pt]
\end{tabular}
\caption{%Trajectory forecasting
Performance on varying detectors. The max recall rate of SSN and PointPillar are lower than 0.8.}
\label{table:detectors}
\vspace{-0.2cm}
\end{table}
\textbf{Qualitative visualization.} Predicted examples are shown in Fig.\ref{fig:qualitative}. %, where a static and a moving objects are detected.
Tracked trajectory in Fig.\ref{fig:qualitative}(a) breaks into two fragments due to the inaccurate detection and affinity measuring, while there is no such problem in our model (Fig.\ref{fig:qualitative}(b)) own to ASU and MSA, which have incorporated the historical association uncertainty into the motion encoding process. As a further observation, since detection results could have FPs caused by insufficient non-maximum suppression, existing methods can hardly distinguish these FPs from TPs when using detection results as input for trajectory forecasting, which may cause inaccurate predicted trajectories for those FPs due to the absence of the historical information (upper dashed blue line in Fig.\ref{fig:qualitative}(a)). In our model, these FPs at the current time can leverage the motion encodings in their vicinity assisted by ASU, and be predicted with future trajectories that are very similar to the future trajectories of the TPs (Fig.\ref{fig:qualitative}(b)).
This feature would be helpful for downstream tasks like motion planning.
\subsection{Comparison with the Existing Studies}\label{sec:comparison}
Quantitative comparison with the existing studies is presented
in Tab.\ref{table:sota}. Two settings are used to keep consistent with the other works:
Setting 1: evaluating at 10Hz with the forecasting error at interpolated frames also considered\cite{liang2020pnpnet}. Setting 2, only the error at the official annotated key frames are considered.
In addition, we report results for both Car and Vehicle in Tab.\ref{table:sota} to match with the other works.
As our model requires detection results, we report the performances of our approach based on two detectors, Centorpoint\cite{yin2021center} and FreeAnchor\cite{zhang2019freeanchor}. SIM function of \cite{zhang2020social} is adopted to our models according to Tab.\ref{table:SIM}.

We first compare the proposed model with PTP\cite{weng2021ptp}, a method also based on detections and towards joint optimization of tracking and trajectory forecasting. In our re-implementation version \footnote{Official code is not available.}, we replace the model with a deterministic forecasting head and evaluate the forecasting based on its online tracking results for a fair comparison.
Although PTP employs GNN to model the interaction between tracks and detections, it still requires explicit historical trajectories up to $t-1$ as input. Our approach does not require any explicitly tracked trajectories and incorporates the association uncertainty into the motion encoding process, which makes it superior performances.

As an extension to the joint detection and forecasting methods, PnPNet\cite{liang2020pnpnet} introduces the tracking in the loop, which achieves the SOTA performance towards forecasting from sensor data. However, PnPNet together with other lidar-based multi-task perception methods\cite{zeng2020dsdnet,li2020end,meyer2020laserflow,casas2020spagnn} are based on their private detectors, which may not be directly comparable with ours as the detector backbones could vary. We list the results of them in Tab.\ref{table:sota} to show the great competitiveness of our proposed method.
Moreover, the proposed uncertain-aware motion encoding is complementary to these studies as it represents the historical information with association uncertainty, which does not conflict with the design of detector backbones or headers. \par
\begin{table}[]
\centering
\small
\begin{tabular}{c|c|l|cc}
\toprule[1pt]
\centering
\multirow{2}{*}{Setting} &\multirow{2}{*}{Class}&\multirow{2}{*}{Approach}  & \multicolumn{2}{c}{fde@3s  $\downarrow$}\\
 &&& rec@0.6& \multicolumn{1}{c}{rec@0.8} \\
\midrule[0.5pt]
\multirow{3}{*}{1}&\multirow{3}{*}{Car}&\cite{yin2021center}+PTP\cite{weng2021ptp}& 122.7 & \multicolumn{1}{c}{126.4}     \\
&&PnPNet\cite{liang2020pnpnet}& 96& \multicolumn{1}{c}{107}     \\

&&\textbf{\cite{yin2021center}+Ours}& {97.7}  & \multicolumn{1}{c}{\textbf{99.9}}     \\
&&\textbf{\cite{zhang2019freeanchor}+Ours}& \textbf{93.4}  & \multicolumn{1}{c}{{105.4}}     \\

\midrule[0.5pt]
\multirow{4}{*}{2}&\multirow{4}{*}{Car}&CAR-Net\cite{sadeghian2018car}  & 158  & \multicolumn{1}{c}{-}     \\
\multirow{3}{*}{}&&NMP\cite{zeng2019end} & 140 & \multicolumn{1}{c}{-}     \\
&&DSDNet\cite{zeng2020dsdnet} & 127 & \multicolumn{1}{c}{-}     \\
&&\textbf{\cite{yin2021center}+Ours} & {106.0} & \multicolumn{1}{c}{\textbf{109.2}}     \\
&&\textbf{\cite{zhang2019freeanchor}+Ours} & \textbf{101.0} & \multicolumn{1}{c}{{113.6}}     \\
\midrule[0.5pt]

\multirow{4}{*}{2}&\multirow{4}{*}{Vehicle}&SPAGNN\cite{casas2020spagnn}& 145  & \multicolumn{1}{c}{-}     \\
&&LaserFlow\cite{meyer2020laserflow}& 143 & \multicolumn{1}{c}{-}     \\
&&CPP-IT\cite{li2020end} & 112.4& \multicolumn{1}{c}{117.9}     \\
%&&LiRaNet &\textbf{102} & \multicolumn{1}{c}{{106}}     \\
&&\textbf{\cite{yin2021center}+Ours} &{105.1} & \multicolumn{1}{c}{\textbf{108.6}}     \\
&&\textbf{\cite{zhang2019freeanchor}+Ours} &\textbf{104.1} & \multicolumn{1}{c}{{114.9}}     \\
\bottomrule[1pt]
\end{tabular}
\caption{Comparison with existing approaches on the nuScenes dataset. Setting 1 includes the error on the interpolated frames\cite{liang2020pnpnet}, setting 2 only considers the error on official key frames.}
\label{table:sota}
\vspace{-0.2cm}
\end{table}

\section{Conclusion}
In this paper, we propose a trajectory predictor based on detection results without prior tracking information. Different from previous works, we incorporate the association uncertainty into the motion encoding process towards trajectory forecasting
through two mechanisms, affinity-aware state update and multiple state aggregation, which alleviate the impact of detection and tracking noise on forecasting results. The proposed model has well generalization ability to adapt to different detectors and advanced trajectory forecasting designs. In the future, we plan to further improve the robustness of the trajectory representation and leverage the forecasting results to boost the detectors.

{\small
\bibliographystyle{ieee_fullname}
\bibliography{script}
}
\end{document}